\documentclass[sigconf, screen, nonacm]{acmart}

\renewcommand\footnotetextcopyrightpermission[1]{}
\AtBeginDocument{%
  }

\usepackage{graphicx} 
\usepackage{xspace}
\usepackage{multirow}
\usepackage{booktabs}
\usepackage{xcolor}
\usepackage[table]{xcolor}  
\usepackage{array}
\usepackage{makecell}
\usepackage{enumerate}
\usepackage{tcolorbox}
\usepackage{listings}
\usepackage{pifont}
\definecolor{skyblue}{RGB}{135,206,235}

\newcommand{\projecttitle}{EvolveNav\xspace}

\newcommand{\revise}[1]{#1}

\newcolumntype{x}[1]{>{\centering\arraybackslash}p{#1pt}}
\newcolumntype{y}[1]{>{\raggedright\arraybackslash}p{#1pt}}
\newcolumntype{z}[1]{>{\raggedleft\arraybackslash}p{#1pt}}

\begin{document}
\title{\projecttitle: Proactive Preflection and Self-Evolving Memory for Zero-Shot Object Goal Navigation}

\settopmatter{
  authorsperrow=4,
  printacmref=false,
  printccs=false
}

\author{Qi Chai}
\authornote{Equal Contribution.}
\affiliation{%
  \institution{HKUST(GZ)}
  \city{Guangzhou}
  \country{China}
}
\email{qchai315@connect.hkust-gz.edu.cn}

\author{Wenhao Shen}
\authornotemark[1]
\affiliation{%
  \institution{Nanyang Technological University}
  \city{Singapore}
  \country{Singapore}
}
\email{wenhao005@e.ntu.edu.sg}

\author{Nanjie Yao}
\affiliation{%
  \institution{HKUST(GZ)}
  \city{Guangzhou}
  \country{China}
}
\email{nanjieyao@gmail.com}

\author{Yue Xia}
\affiliation{%
  \institution{Xi'an Jiaotong University}
  \city{Xi'an}
  \country{China}
}
\email{xiayue@stu.xjtu.edu.cn}

\author{Kaiyong Zhao}
\affiliation{%
  \institution{XGRIDS}
  \city{Shenzhen}
  \country{China}
} 
\email{kyzhao@xgrids.com}

\author{Jie Ma}
\affiliation{%
  \institution{Xi'an Jiaotong University}
  \city{Xi'an}
  \country{China}
}
\email{jiema@xjtu.edu.cn}

\author{Guosheng Lin}
\affiliation{%
  \institution{Nanyang Technological University}
  \city{Singapore}
  \country{Singapore}
}
\email{gslin@ntu.edu.sg}

\author{Hao Wang}
\authornote{Corresponding author.}
\affiliation{%
  \institution{HKUST(GZ)}
  \city{Guangzhou}
  \country{China}
}
\email{haowang@hkust-gz.edu.cn}


\begin{abstract}
Zero-Shot Object-Goal Navigation (ZS-OGN) requires embodied agents to explore and locate target objects without any prior training. To this end, recent methods leverage foundation models. But they typically rely on static priors and lack adaptation, which leads to repeated errors and costly trial and error.
In this paper, we propose a self-evolving ZS-OGN framework that enables continuous test-time improvement. Specifically, we build an agentic rule memory by extracting actionable knowledge from past trajectories. Then, we propose a retrieval strategy based on upper confidence bound, selecting effective rules by balancing semantic relevance and historical success. In addition, we introduce a memory-guided preflection module that forecasts potential outcomes before action, reducing inefficient exploration.
Extensive experiments show that our method outperforms existing zero-shot baselines, achieving a 10.1\% improvement in success rate with fewer unnecessary steps. 
\end{abstract}



\begin{teaserfigure}
  \centering
  \includegraphics[width=0.99\textwidth]{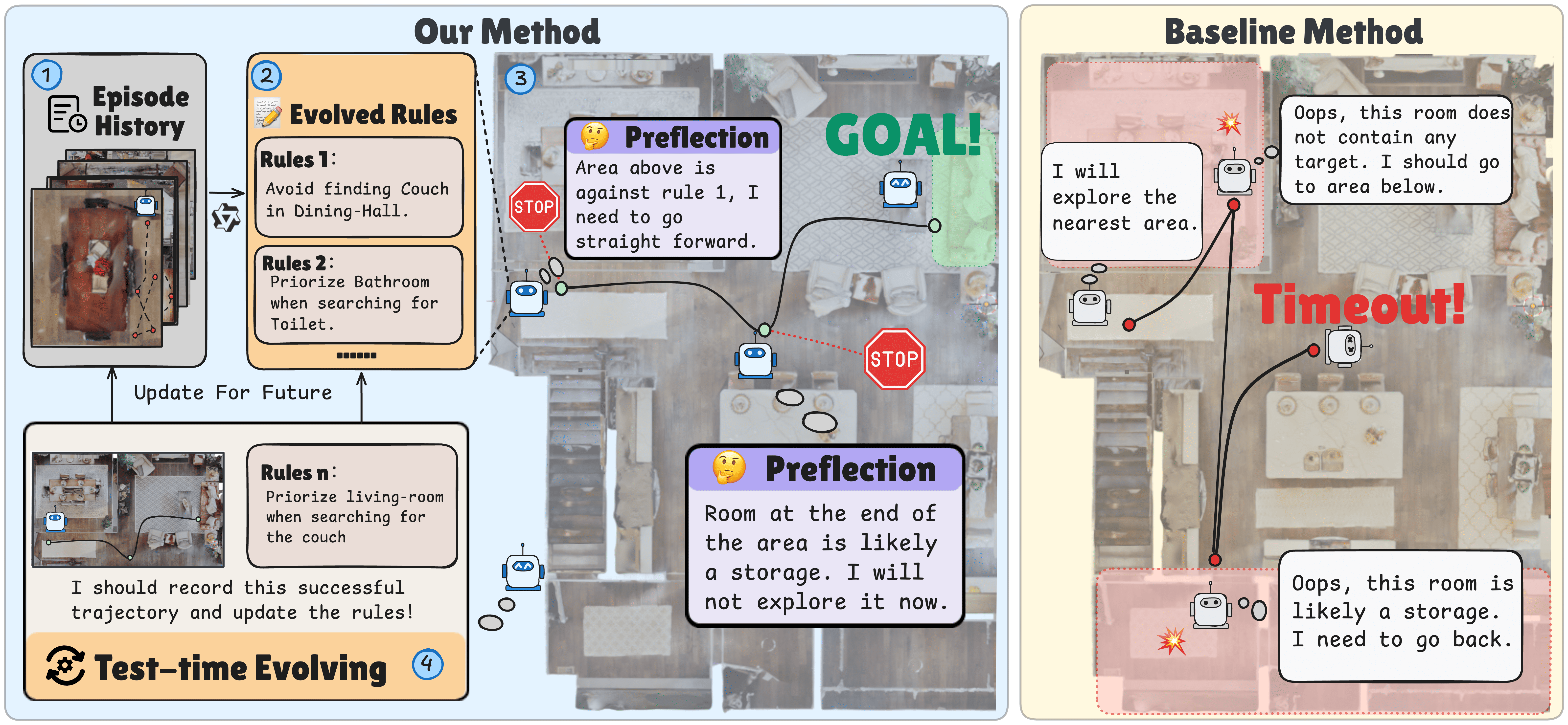}
  \vspace{-0.3cm}
  \caption{{Method Comparison. \textit{Left}: Our agent learns from (1) past episode history to dynamically update (2) evolved rules. During navigation, it adopts (3) preflection to proactively avoid rule-violating or unpromising rooms, while continuously (4) evolving its rule set from trajectories. \textit{Right}: Baseline method relies on passive reaction, leading to inefficient exploration.}}
  \label{fig:teaser}
\end{teaserfigure}


\newcommand{\cq}[1]{\textcolor{purple}{#1}}
\newcommand{\ynj}[1]{\textcolor{blue}{#1}}
\newcommand{\swh}[1]{\textcolor{red}{#1}}

\maketitle

\section{Introduction}

Object-Goal Navigation (OGN) is a fundamental task in embodied artificial intelligence~\cite{sun2024comprehensive, savva2019habitat}. It requires an agent to perceive, reason, and act within an unseen space and locate a specific target. 
Existing OGN methodologies fall into training-based and training-free paradigms. Training-based methods~\cite{chen2022think, tantrue, ramrakhya2023pirlnav, dang2023search, wijmans2019dd, wasserman2024exploitation}, relying on Reinforcement Learning (RL) or Imitation Learning (IL), demand massive datasets or extensive simulation. This leads to sample inefficiency and poor generalization in unseen environments.  

To address this, recent training-free methods~\cite{zhang2025apexnav, gong2026ascent, cai2024bridgingpixnav, yangefficientnav} leverage foundation models like Large Language Models and Vision-Language Models (LLMs/VLMs) as semantic planners.
However, these foundation models only provide general and fixed semantic priors. As a result, Zero-Shot OGN (ZS-OGN) agents cannot adapt to real-time feedback and often repeat identical mistakes, e.g., being consistently misled by visually similar objects. 

Another key limitation is the high cost of trial and error. Under limited step budgets, each action is expensive. Existing methods often rely on post-hoc correction~\cite{kim2023context, wang-etal-2024-e2cl}, where errors are detected only after execution, leading to wasted steps. To improve efficiency, agents must anticipate outcomes before acting and avoid low-value decisions in advance.

Therefore, this paper aims to resolve a critical challenge in OGN: \textit{how to jointly achieve online adaptation and action anticipation, under limited test-time step budgets.}

Specifically, we propose a self-evolving and training-free OGN framework \projecttitle. The key idea is to transform navigation into a continuous self-improvement process, where the agent learns actionable navigation rules from past successes and failures. To this end, we introduce a self-evolving agentic rule memory. After each episode, the agent analyzes its trajectory and visual observations. It then extracts concrete rules, such as recognizing misleading layouts or identifying effective search patterns. This memory is updated across test-time episodes in a dynamic manner, which ensures the accumulated knowledge remains relevant and reliable.


To utilize this knowledge during exploration, we design a rule retrieval strategy based on Upper Confidence Bound (UCB). Each rule is scored by balancing two factors: its semantic relevance to the current scene and its historical success rate. This allows the agent to select rules that are both contextually appropriate and empirically reliable. As a result, the agent continuously improves its behavior through experience, by updating the UCB scores only.


In addition, to address the high cost of ZS-OGN agent trial and error, we propose a memory guided preflection module. Instead of correcting mistakes after actions are executed, the agent evaluates candidate waypoints before acting. It retrieves relevant rules and past failure cases to anticipate future observations and potential risks. This process enables the agent to filter out low-value or high-risk actions in advance, leading to more efficient and goal-directed exploration in unseen environments.

Our main contributions are summarized as follows:

\begin{itemize}
    \item We introduce a self-evolving agentic rule memory to support online experience accumulation. With a UCB-based strategy that balances semantic relevance and historical success, the agent continuously refines its behavior.
    
    \item We propose a memory-guided preflection module that shifts the ZS-OGN task from passive correction to proactive risk avoidance. By anticipating potential failures before execution, it effectively mitigates physical trial-and-error costs.
    
    \item Extensive experiments show our method significantly outperforms zero-shot baselines, notably achieving a 10.1\% success rate improvement on MP3D while minimizing redundant exploration.
\end{itemize}

\section{Related Work}
\label{sec: relatedwork}
\subsection{Object Goal Navigation} 

Object goal navigation requires an agent to navigate unknown environments to find specified target objects. Early approaches primarily relied on end-to-end RL~\cite{ye2021efficient} or modular learning~\cite{chaplot2020object}. While RL-based methods directly map observations to actions, they often suffer from poor generalization and sim-to-real gaps. Modular methods mitigate training costs by building semantic maps but still struggle to generalize to unseen environments without task-specific training. 

Recently, LLMs~\cite{singh2025openaigpt5card, bai2023qwen} and VLMs~\cite{li2023blip} have been introduced to enable zero-shot OGN. Methods like ZSON~\cite{majumdar2022zson}, CoW~\cite{gadre2023cows}, and VLFM~\cite{yokoyama2024vlfm} utilize VLMs to evaluate semantic similarity between frontiers and the target to guide exploration. Meanwhile, approaches like L3MVN~\cite{yu2023l3mvn} and SG-Nav~\cite{yin2025sg} leverage LLMs and 3D scene graphs for high-level spatial reasoning and long-horizon planning. To further improve efficiency and robustness in complex real-world scenarios, recent works have made significant advancements. ApexNav~\cite{zhang2025apexnav} introduces an adaptive exploration strategy balancing semantic and geometric exploration, along with a target-centric semantic fusion to reduce visual misdetections of similar objects. MSGNav~\cite{huang2025msgnav} constructs a multi-modal 3D scene graph that retains images as edge features, preventing visual information loss and enhancing robust open-vocabulary reasoning. To address the single-floor limitation, ASCENT~\cite{gong2026ascent} proposes an online cross-floor navigation framework featuring a hierarchical stair-aware topology and an LLM-driven coarse-to-fine reasoning module for efficient multi-floor exploration. 
However, these methods rely on fixed priors during deployment and struggle to adapt to real-time feedback, often repeating mistakes. 


\begin{figure*}
    \centering
    \includegraphics[width=1.0\linewidth]{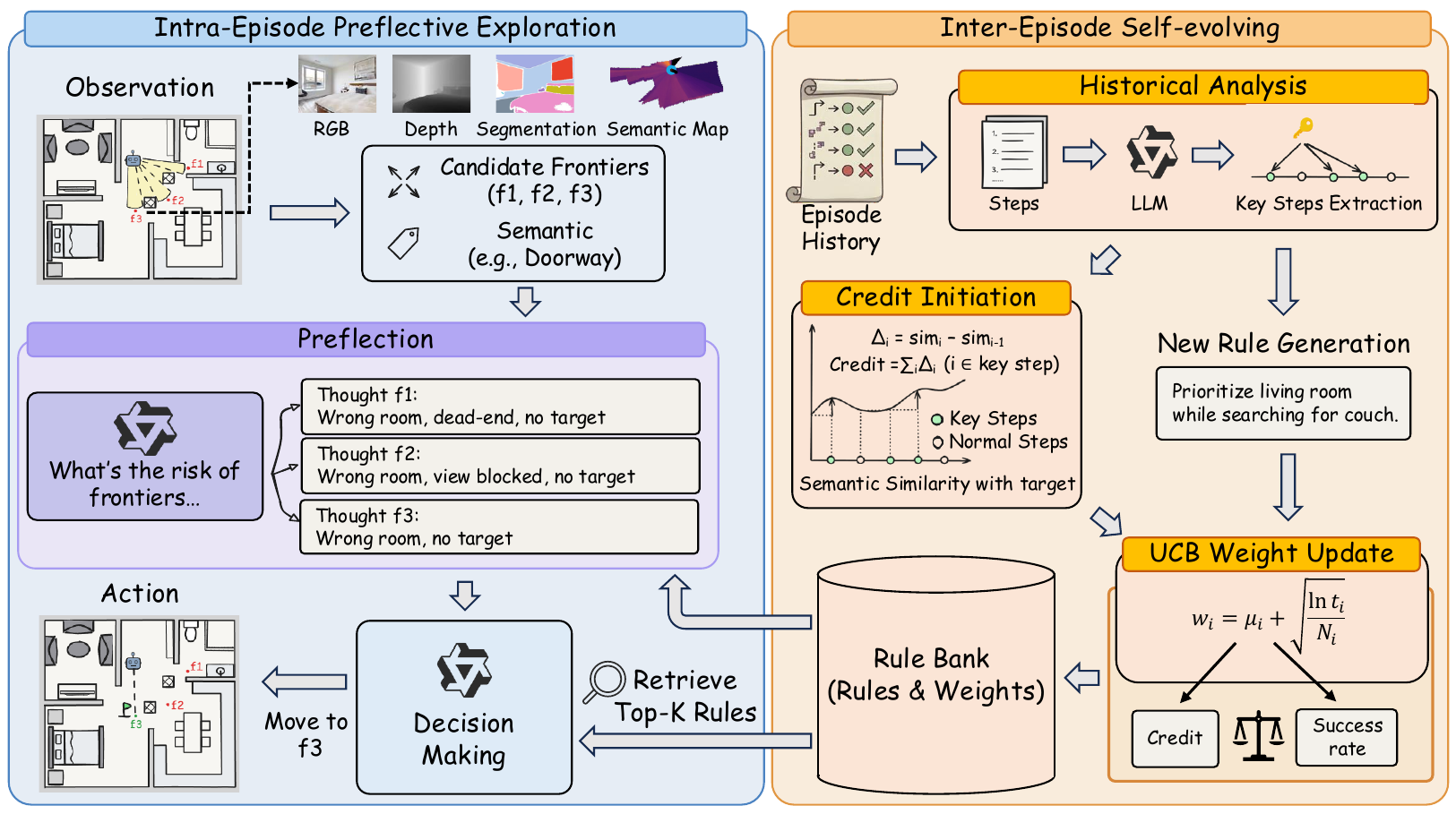}
    \vspace{-0.4cm}
    \caption{
    Method Overview of \projecttitle. The architecture consists of two core components for continuous optimization. \emph{\textbf{Left (Intra-Episode Preflective Exploration):}} During test-time execution, the agent evaluates frontiers ($f_1, f_2, f_3, \ldots, f_n$) by retrieving top-$K$ rules from the rule bank for LLM-driven preflection. 
    This assesses risks to determine the optimal direction, minimizing trial-and-error.
    \emph{\textbf{Right (Inter-Episode Rule Self-Evolution):}} After each navigation episode, the framework distills new rules from the trajectory history. A UCB-based algorithm updates rule weights by balancing credit assignment and exploration potential, continuously refining the rule bank.
    }
    \label{fig: pipeline}
    \vspace{-0.4cm}
\end{figure*}

\subsection{Self-evolution in Agents} 
LLM agents have demonstrated strong capabilities across a wide range of domains~\cite{fu2025vistawise, chai2025causalmace, zhang2025dvm, ma2025debate}. In recent years, investigating the capacity for self-evolution within autonomous agents has garnered significant attention in recent research. 
Existing self-evolving techniques can be broadly divided into two main areas: prompt optimization and memory optimization. For prompt optimization, methods~\cite{zhang2025enhancing, zhang2025multimind, wang2023promptagent, yao2023retroformer, xiang2025self, wu2024strago} aim to automatically search for high-quality instructions to guide language models toward producing more accurate outputs. For instance, PromptAgent~\cite{wang2023promptagent} formulates prompt optimization as a strategic planning problem and leverages Monte Carlo tree search to efficiently navigate the expert-level prompt space. Similarly, OPRO~\cite{yang2023large} generates new instructions by prompting the model with previously generated candidates to iteratively refine prompt quality. 

On the other hand, memory optimization methods~\cite{zhong2024memorybank, chhikara2025mem0, wang2024agent, chen2025compress} focus on managing contextual information and past experiences to support long-term decision-making and consistency. Reflexion~\cite{shinn2023reflexion} enables agents to reflect on task feedback and store episodic insights, which promotes continuous self-improvement over time without weight updates. Concurrently, MemoryBank~\cite{zhong2024memorybank} hierarchically summarizes events and dynamically updates memory based on recency and relevance to mitigate forgetting and enhance long-term retention. 
Despite these advancements, self-evolution remains under-explored in OGN. Current frameworks lack the capacity to internalize exploration experiences, causing agents to repeat navigational errors with limited generalization.



\section{Methodology}
\subsection{Problem Formulation}
In this paper, we focus on the zero-shot object-goal navigation (ZS-OGN) task. Unlike traditional learning-based paradigms that rely on massive environment interactions to learn navigation policies, the training-free paradigm requires the agent to navigate in unstructured environments and locate an instance of a specific target object category $c$ (e.g., ``couch''), without any fine-tuning.

Formally, we follow the common practice \cite{zhang2025apexnav,huang2025msgnav,gong2026ascent} for the ZS-OGN task evaluation, an episode begins with the agent placed at an initial pose $p_0$ within an unexplored scene. At any given time step $t$, the agent captures an egocentric RGB-D image $I_t = \langle I_t^{rgb}, I_t^{depth} \rangle$ and obtains its spatial tracking pose $p_t = (x_t, y_t, \theta_t)$. Relying on these sensory inputs alongside its maintained memory state, the agent determines and performs a discrete action $a_t \in \mathcal{A}$. Specifically, the permissible action space $\mathcal{A}$ is defined as a set of six distinct commands: \texttt{MOVE\_FORWARD}, \texttt{TURN\_LEFT}, \texttt{TURN\_RIGHT}, \texttt{LOOK\_UP}, \texttt{LOOK\_DOWN}, and \texttt{STOP}.
A navigation episode ends when the agent executes the \texttt{STOP} action or exceeds the maximum step limit $T$. The task succeeds if the agent stops within $1\text{ meter}$ of the visible target object $c$. Otherwise, the episode is considered a failure.

\subsection{Overview}
In time-constrained training-free navigation, an agent must minimize inefficient physical trial and error and continuously self-improve by leveraging historical successes and failures. To achieve this without any environment-specific parameter fine-tuning, we propose a novel navigation framework that tightly couples real-time spatial exploration with a self-evolving cognitive memory. As illustrated in Figure~\ref{fig: pipeline}, our system is composed of two primary components: Intra-Episode Preflective Exploration and Inter-Episode Rule Self-Evolution:

\noindent $\bullet$ \textbf{Intra-Episode Preflective Exploration.} Within a navigation episode, the agent actively constructs a 2D environment map and extracts candidate frontiers. Before moving, an LLM retrieves past empirical rules from our constructed memory. It then performs ``preflection'' to evaluate the potential risks of each frontier and select the optimal exploration direction. (Section~\ref{sec:preflection})

\noindent $\bullet$ \textbf{Inter-Episode Rule Self-Evolution.} After each episode, the agent performs a post-hoc analysis to enable continuous learning. It uses the LLM to review the trajectory, identify key decisions, and distill them into generalizable navigation rules. To manage this growing rule bank, we apply a UCB algorithm to dynamically balance the exploitation of proven rules and the exploration of new ones (Section~\ref{sec:memory}).



\subsection{Spatial Mapping and Low-Level Control}
\label{sec:mapping}
To effectively navigate in unknown and unstructured environments, the agent must first establish a geometric understanding of the surrounding physical space and possess reliable low-level locomotion capabilities. 
At each time step $t$, the agent leverages egocentric RGB-D observations and odometry to incrementally project 3D depth data into a global coordinate frame, dynamically maintaining a 2D Occupancy Grid Map. 
By mapping 3D data onto a 2D plane, this projection reduces computational complexity while retaining critical information regarding traversable areas and obstacle positioning.

Based on this occupancy map, the system continuously extracts a set of candidate frontiers. A frontier is defined as the geometric boundary separating known free space from unexplored regions. By explicitly identifying these frontiers, the agent can systematically target the most informative edges of its current knowledge, ensuring that exploration is driven by spatial uncertainty rather than random wandering.

Once a target coordinate is determined by the LLM planner (detailed in Section~\ref{sec:preflection}) or identified as the target object $c$ by a pre-trained vision model, it becomes a 2D local waypoint. A low-level navigation controller~\cite{ZhaoICCV2021pointnav} then navigates the agent to this waypoint. 
This hierarchical design enhances system efficiency by selectively invoking the LLM only when necessary, thereby minimizing redundant API calls.
To prevent the agent from getting stuck, the controller executes movement for a maximum of $H$ steps. Execution stops early if the frontier is fully explored, the waypoint becomes unreachable, or a visual trigger starts a new decision cycle. This step bound ensures the agent stays reactive to new observations and recovers quickly from failures.

\subsection{Self-evolving Agentic Rule Memory}
\label{sec:memory}
In zero-shot navigation, agents lack the ability to update their model parameters, making them highly susceptible to repeating the same mistakes across different environments. 
To achieve continuous self-improvement without fine-tuning, our framework introduces a post-episode reflection mechanism that distills low-level trajectory data into highly generalizable cognitive rules. Regardless of whether an episode ends in success or failure, the system compiles the agent's complete history of frontier decisions and executes a two-stage rule generation pipeline. 
First, the LLM reviews the global trajectory to isolate the pivotal decision steps that determined the episode's outcome. 
Subsequently, a secondary LLM query analyzes the local observations and decision contexts of these specific steps to synthesize reusable navigation rules. 
Finally, these newly formulated rules are deposited into the rule memory bank $\mathcal{M}$ for future invocation and dynamic weight evaluation.


\subsubsection{Semantic-driven Credit Assignment} 
Distilling high-quality rules from a lengthy exploration trajectory requires identifying exactly which actions led to the final outcome. We must therefore assign accurate ``credit'' to each step, mathematically quantifying its specific contribution and importance. Because a single episode can encompass dozens of steps, simply treating all actions equally is ineffective. Inspired by information potential theory~\cite{xietips}, we propose a credit assignment mechanism based on sequential improvements in semantic similarity. For the $i$-th step in the trajectory, we calculate a composite semantic similarity score $\mathcal{S}_i$ between the currently observed scene and the target object, reflecting the potential value of that step:
\begin{equation}
    \mathcal{S}_i = \alpha \cdot \mathcal{S}_{\text{room}}^{(i)} + (1-\alpha) \cdot \mathcal{S}_{\text{obj}}^{(i)},
\end{equation}
where $\mathcal{S}_{\text{room}}^{(i)}$ and $\mathcal{S}_{\text{obj}}^{(i)}$ represent the cosine similarities of the currently observed room type and the visible object set against the target category, respectively, and $\alpha$ serves as a balancing coefficient. The initial credit $C_i$ assigned to the $i$-th step is then defined as the marginal gain in semantic similarity relative to the preceding step:
\begin{equation}
    C_i = \mathcal{S}_i - \mathcal{S}_{i-1}.
\end{equation}

To differentiate the reward distribution between successful and failed episodes, we apply a clipping operation to $C_i$ to prevent extreme values and compute a normalized weight $w_i$ using a softmax function. Ultimately, the initial support score $S_r$ assigned to the newly generated rule $r$ is defined as the weighted sum of the contributions from its associated key steps:
\begin{equation}
    S_r = \sum_{i} w_i \cdot C_i.
\end{equation}

\subsubsection{Dynamic Memory Management via UCB} 
As the agent continuously explores multiple environments, the rule memory bank naturally expands. Given the prohibitive physical trial-and-error costs associated with training-free navigation, the system must rigorously balance the exploitation of historically proven rules with the exploration of newly formulated, unverified ones. To mathematically formalize this trade-off, we treat rule retrieval as a multi-armed bandit problem and employ a UCB management strategy.

For any rule $r \in \mathcal{M}$, its expected utility $\mu_r$ combines historical performance with the latest episodic feedback. This utility is iteratively smoothed using a momentum-based update to incorporate the newly computed support score $S_r$:
\begin{equation}
    \mu_r \leftarrow \eta \cdot \mu_r + (1 - \eta) \cdot S_r,
\end{equation}
where $\eta \in [0,1]$ serves as a momentum decay coefficient. Prior to initiating a new navigation episode, the framework computes the UCB score for all rules residing in the memory bank:
\begin{equation}
    \text{UCB}(r) = \mu_r + \beta \sqrt{\frac{\ln T}{N_r}},
\end{equation}
where $T$ represents the total number of exploration episodes the agent has completed thus far, $N_r$ is the historical frequency with which rule $r$ has been retrieved and applied, and $\beta$ is a hyperparameter modulating the exploration degree. 

Crucially, to ensure that newly generated rules receive a guaranteed initial validation phase, their UCB scores are initialized to infinity ($\infty$) when $N_r = 0$. Finally, the memory bank is sorted in descending order based on these computed $\text{UCB}(r)$ scores. 
The top-$K$ rules are extracted and injected as experiential priors into the LLM's context during the preflection phase (as detailed in Section~\ref{sec:preflection}), thereby successfully completing the self-improving cognitive loop of the framework.

\subsection{Memory-Guided Preflection}
\label{sec:preflection}


As established in Section~\ref{sec:mapping}, the mapping module identifies a set of candidate frontiers at each decision step. In environments with constrained step budgets, relying on greedy distance-based strategies or static commonsense priors often traps the agent in local optima or leads to repetitive exploration of irrelevant regions. To circumvent these prohibitively expensive physical trial-and-error costs, we propose an LLM-based memory-guided preflection mechanism. This module enables the agent to proactively anticipate and avoid potential failures before executing irreversible physical actions.

To implement this, we construct a structured context prompt that integrates geometric distances with local semantic observations, such as visible objects and spatial layouts for each candidate. Unlike traditional training-free methods that rely exclusively on fixed commonsense priors, which often struggle with atypical environments, our approach dynamically injects the agent’s self-summarized decision memory into this reasoning context

Specifically, at each decision stage, the system queries a self-evolving rule bank $\mathcal{M}$ (detailed in Section~\ref{sec:memory}) to retrieve the top-$K$ most relevant experiential rules. To balance the exploitation of reliable past rules with the exploration of newly formed insights, the retrieval is governed by an Upper Confidence Bound (UCB) selection criterion:
\begin{equation}
    \mathcal{R}_{\text{top}K} = \operatorname{TopK}_{r \in \mathcal{M}} \big( \text{UCB}(r) \big),
\end{equation}
where $\text{UCB}(r)$ denotes the confidence score of rule $r$. By grounding the prompt in these dynamically retrieved rules, the agent leverages its own historical lessons to avoid repeating systemic mistakes.

To mitigate hallucinations and impulsive decisions, we replace direct target selection with an explicit reasoning process. The LLM must first perform failure prediction for each candidate frontier by synthesizing injected rules with current observations. For example, it might predict a high risk of failure if a frontier's visual cues conflict with established spatial rules. This mandatory risk assessment filters out deceptive paths, allowing the LLM to balance failure probabilities against exploration rewards to select the optimal, risk-averse frontier.

\section{Experiment}

\subsection{Experimental Setup}
\subsubsection{Datasets} We evaluate our proposed method on two standard ObjectNav benchmarks derived from the Habitat Challenge. Specifically, we utilize the HM3D-Semantics-v0.1 dataset from the 2022 Habitat challenge~\cite{yadav2023habitat}, which comprises 2,000 evaluation episodes across 20 distinct scenes and 6
goal categories. Additionally, we evaluate on the updated MP3D dataset from the 2021 Habitat challenge~\cite{chang2017matterport3d}, consisting of 2195 episodes, 11 scenes, 21 goal categories episodes distributed over the 11 scenes. 

\subsubsection{Evaluation Metrics} Following the standard evaluation protocol established in prior ObjectNav methods~\cite{zhang2025apexnav}, we employ Success Rate (SR) and Success weighted by Path Length (SPL) to comprehensively evaluate the performance of our proposed method. SR measures the fundamental navigation efficacy, defined as the fraction of episodes where the agent successfully navigates to the target object and executes the \textit{stop} action. SPL further assesses navigation efficiency by weighting the binary success indicator of each episode against the ratio of the shortest path distance to the actual agent trajectory length, thereby penalizing circuitous routes.

\subsubsection{Baselines} To evaluate the effectiveness of our proposed approach, we compare it against a comprehensive set of state-of-the-art baselines, which are broadly categorized into learning-based and zero-shot methods. The learning-based category includes RIM~\cite{chen2023object}, OVG-Nav~\cite{yoo2024commonsense}, PIRLNav~\cite{ramrakhya2023pirlnav}, and XGX~\cite{wasserman2024exploitation}, which heavily rely on task-specific training architectures to achieve navigation. In the zero-shot category, we evaluate foundational methods such as ZSON~\cite{majumdar2022zson} and ActPept~\cite{guo2024object}, which utilize contrastive models like CLIP or graph-based structures for semantic guidance, alongside progressive frameworks that incorporate LLMs and VLMs for explicit instruction interpolation. Specifically, we compare against L3MVN~\cite{yu2023l3mvn}, PixNav~\cite{cai2024bridgingpixnav}, SG-Nav~\cite{yin2025sg}, OpenFMNav~\cite{kuang2024openfmnav}, and InstructNav~\cite{longinstructnav}, which integrate various iterations of GPT architectures to interpret language and guide exploration. Furthermore, we benchmark against recent advanced zero-shot frameworks including VLFM~\cite{yokoyama2024vlfm} and ApexNav~\cite{zhang2025apexnav}, which employ BLIP-2~\cite{li2023blip} paired with robust models like DeepSeek-V3~\cite{liu2024deepseek}, as well as the latest state-of-the-art methods such as MFNP~\cite{zhang2024multi}, ASCENT~\cite{gong2026ascent}, and MSGNav~\cite{huang2025msgnav}, which leverage the Qwen model family~\cite{bai2023qwen, yang2024qwen2,yang2024qwen25,yang2025qwen3} to construct reliable visual-linguistic mappings for complex navigation tasks.

\subsubsection{Implementation Details} For the multimodal reasoning backbone, we employ the pre-trained BLIP-2~\cite{li2023blip} as our vision foundation model and Qwen3-8B~\cite{yang2025qwen3} as the language model instruction interpolator. In strict adherence to the zero-shot object-goal navigation paradigm, all network parameters remain entirely frozen during deployment, with no task-specific or environment-specific fine-tuning applied. The core functional components, namely the memory-guided preflection module and the self-evolving agentic rule memory, operate purely at inference time through dynamic prompt injection and UCB-based rule updating. Following established evaluation protocols from prior works~\cite{gong2026ascent, huang2025msgnav, batra2020objectnav}, we set the success distance thresholds to 1.0 meters for both benchmarks. Regarding the computational infrastructure, all simulated experiments, environment rendering, and large model inference processes are executed on a single high-performance workstation equipped with a AMD EPYC 7542 32-Core Processor and 2 NVIDIA RTX A6000 GPUs. Additional hyperparameter configurations and further implementation details are provided in the Supplementary Material.

\subsection{Quantitative Evaluation}

\begin{table*}[t]
    \centering
    \caption{{\textbf{Comparisons with SOTA methods.} 
The table compares learning-based and zero-shot methods on HM3D and MP3D datasets across the \textit{``Success Rate''} (SR) and \textit{``Success Weighted by Path Length''} (SPL) metrics. We introduce columns for \textit{Vision} and \textit{Language} models to specify the Instruction Interpolator components used by each approach. All baseline results from their original publications. Arrow ``$\uparrow$'' denotes higher is better. Our method achieves state-of-the-art performance among zero-shot approaches, yielding the highest SR and SPL on both benchmarks.}}
    \begin{tabular}{x{65}|x{60}|x{40}|x{90}x{75}|cc|cc}
        \toprule
        
         \multirow{2.5}{*}{\textbf{Method}}  & \multirow{2.5}{*}{\textbf{Venue}} & \multirow{2.5}{*}{\textbf{Zero-shot}} & \multicolumn{2}{c|}{\textbf{Instruction Interpolator}} & \multicolumn{2}{c|}{\textbf{HM3D} \cite{ramakrishnan2021habitat}} & \multicolumn{2}{c}{\textbf{MP3D} \cite{chang2017matterport3d}}
        \\
        \cmidrule(lr){4-5} \cmidrule(lr){6-7} \cmidrule(lr){8-9}
         & & & \textbf{Vision} & \textbf{Language}  & \textbf{SR} $\uparrow$ & \textbf{SPL} $\uparrow$ & \textbf{SR} $\uparrow$ & \textbf{SPL} $\uparrow$  
        \\
        \midrule
         \textcolor{gray}{RIM \cite{chen2023object}}  & {\small \textcolor{gray}{IROS'23}} & \textcolor{gray}{\ding{55}}  & \textcolor{gray}{-} & \textcolor{gray}{-} & \textcolor{gray}{$57.8$} & \textcolor{gray}{$27.2$} & \textcolor{gray}{$50.3$} & \textcolor{gray}{$17.0$} 
        \\
         \textcolor{gray}{OVG-Nav \cite{yoo2024commonsense}}  & {\small \textcolor{gray}{RAL'24}} & \textcolor{gray}{\ding{55}}  & \textcolor{gray}{-} & \textcolor{gray}{-} & \textcolor{gray}{-} & \textcolor{gray}{-} & \textcolor{gray}{$35.8$} & \textcolor{gray}{$12.3$} 
        \\
         \textcolor{gray}{VLFM \cite{yokoyama2024vlfm}}  & {\small \textcolor{gray}{ICRA'24}} & \textcolor{gray}{\ding{55}}  & \textcolor{gray}{BLIP-2~\cite{li2023blip}}& - & \textcolor{gray}{$52.5$} & \textcolor{gray}{$30.4$} & \textcolor{gray}{$36.4$} & \textcolor{gray}{$17.5$}
         \\
         \textcolor{gray}{PIRLNav \cite{ramrakhya2023pirlnav}} & {\small \textcolor{gray}{CVPR'23}} & \textcolor{gray}{\ding{55}}  & \textcolor{gray}{-} & \textcolor{gray}{-} & \textcolor{gray}{$64.1$} & \textcolor{gray}{$27.1$} & \textcolor{gray}{-} & \textcolor{gray}{-} 
        \\
         \textcolor{gray}{XGX \cite{wasserman2024exploitation}}  & {\small \textcolor{gray}{ICRA'24}} & \textcolor{gray}{\ding{55}}  & \textcolor{gray}{-} & \textcolor{gray}{-} & \textcolor{gray}{$72.9$} & \textcolor{gray}{$35.7$} & \textcolor{gray}{-} & \textcolor{gray}{-}
        \\
        \midrule
        ZSON \cite{majumdar2022zson} & {\small NeurIPS'22} & \ding{51}  & CLIP \cite{radford2021learning} & - & $25.5$ & $12.6$ & $15.3$ & $4.8$
        \\
         L3MVN \cite{yu2023l3mvn} & {\small IROS'23}  & \ding{51}  & - & RoBERTa-large \cite{liu2019roberta} & $50.4$ & $23.1$ & $34.9$ & $14.5$
        \\  
         PixNav \cite{cai2024bridgingpixnav} & {\small ICRA'24} & \ding{51}  & LLaMA-Adapter \cite{zhang2023llama} & \raisebox{-0.2\height}{\includegraphics[width=0.075\linewidth]{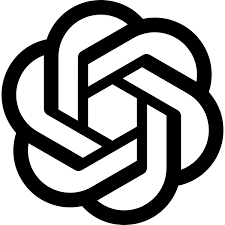}} GPT-4 \cite{achiam2023gpt} & $37.9$ & $20.5$ & - & -
        \\
         VLFM \cite{yokoyama2024vlfm} & {\small ICRA'24} & \ding{51}  & BLIP-2 \cite{li2023blip} & - & $50.9$ & $23.6$ & $32.5$ & $15.9$
        \\
         SG-Nav \cite{yin2025sg} & {\small NeurIPS'24} & \ding{51}  & LLaVA-1.6-7B \cite{liu2023visual} & \raisebox{-0.2\height}{\includegraphics[width=0.075\linewidth]{icons/gpt.png}} GPT-4 \cite{achiam2023gpt} & $54.0$ & $24.9$ & $40.2$ & $16.0$
        \\
         \revise{OpenFMNav \cite{kuang2024openfmnav}} & \revise{{\small NAACL-F'24}} & \ding{51} & \revise{\raisebox{-0.2\height}{\includegraphics[width=0.075\linewidth]{icons/gpt.png}} GPT-4V \cite{achiam2023gpt}} & \revise{\raisebox{-0.2\height}{\includegraphics[width=0.075\linewidth]{icons/gpt.png}} GPT-4 \cite{achiam2023gpt}} & \revise{$54.9$} & \revise{$24.4$} & \revise{$37.2$} & \revise{$15.7$}
        \\
         ActPept \cite{guo2024object} & {\small RAL'24} & \ding{51}  & GraphSAGE \cite{hamilton2017inductive} & - & - & - & $39.8$ & $17.4$
        \\
         InstructNav \cite{longinstructnav} & {\small CoRL'24} & \ding{51}  & \raisebox{-0.2\height}{\includegraphics[width=0.075\linewidth]{icons/gpt.png}} GPT-4V \cite{achiam2023gpt} & \raisebox{-0.2\height}{\includegraphics[width=0.075\linewidth]{icons/gpt.png}} GPT-4 \cite{achiam2023gpt} & $58.0$ & $20.9$ & - & - 
        \\
         \revise{ApexNav \cite{zhang2025apexnav}} & \revise{{\small RAL'25}} & \ding{51}  & \revise{BLIP-2 \cite{li2023blip}} & \revise{\raisebox{-0.2\height}{\includegraphics[width=0.11\linewidth]{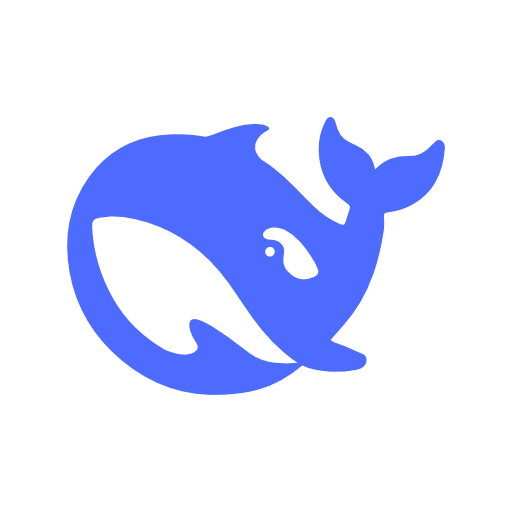}} DeepSeek-V3 \cite{liu2024deepseek}} & \revise{$59.6$} & \revise{$33.0$} & \revise{$39.2$} & \revise{$17.8$}
        \\
        MFNP~\cite{zhang2024multi} & {\small ICRA'25} & \ding{51}  & \raisebox{-0.2\height}{\includegraphics[width=0.06\linewidth]{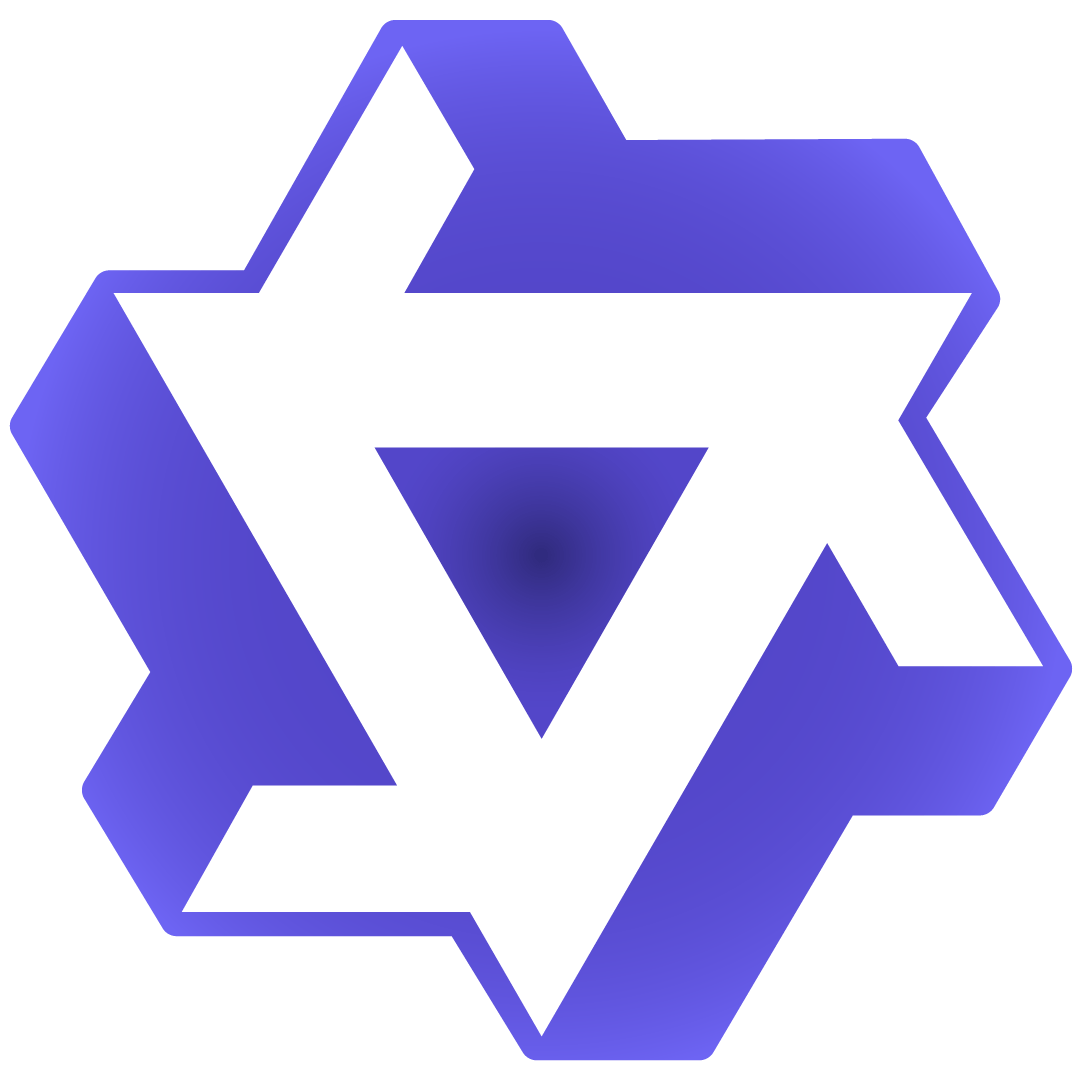}} Qwen-VLChat-Int4 \cite{bai2023qwen} & \raisebox{-0.2\height}{\includegraphics[width=0.085\linewidth]{icons/qwen.png}} Qwen2-7B \cite{yang2024qwen2} & $58.3$ & $26.7$ & $41.1$ & $15.4$
        \\            
        MSGNav~\cite{huang2025msgnav} &  \revise{{\small CVPR'26}} & \ding{51} & \raisebox{-0.2\height}{\includegraphics[width=0.075\linewidth]{icons/gpt.png}} GPT-4o \cite{hurst2024gpt} & \raisebox{-0.2\height}{\includegraphics[width=0.075\linewidth]{icons/gpt.png}} GPT-4o \cite{hurst2024gpt} &  $63.0$ &  $31.4$ &    -& - \\
        ASCENT~\cite{gong2026ascent} & \revise{{\small RAL'26}} & \ding{51} & BLIP-2~\cite{li2023blip}   &\raisebox{-0.2\height}{\includegraphics[width=0.085\linewidth]{icons/qwen.png}} Qwen2.5-7B~\cite{yang2024qwen25} &  $65.4$ &  $33.5$ &    $44.5$ & $15.5$ 
        \\

        \midrule
        \rowcolor{skyblue!40} \textbf{\projecttitle~(Ours)} &  \textbf{-} & \ding{51} & BLIP-2~\cite{li2023blip} & \raisebox{-0.2\height}{\includegraphics[width=0.085\linewidth]{icons/qwen.png}} Qwen3-8B~\cite{yang2025qwen3} &  $\mathbf{67.3}$ &  $\mathbf{33.9}$ &    $\mathbf{49.0}$ & $\mathbf{19.1}$ \\
        \bottomrule
    \end{tabular}
    \label{tab:main_table}
\end{table*}

We compare our proposed approach against state-of-the-art learning-based and zero-shot object goal navigation methods on the widely adopted HM3D and MP3D benchmarks. As demonstrated in Table~\ref{tab:main_table}, our method achieves state-of-the-art performance across both datasets, consistently outperforming existing training-free approaches. On the HM3D dataset, our method sets a new benchmark with an SR of 67.3\% and a Success weighted by SPL of 33.9\%, surpassing recently proposed frameworks such as MSGNav and ASCENT by 1.9\% in SR and 0.4\% in SPL. The performance gains are even more significant on the highly challenging MP3D dataset, which is characterized by larger spatial scales and complex multi-floor layouts. In this environment, our method achieves an impressive 49.0\% SR and 19.1\% SPL, yielding a substantial absolute improvement of 4.5\% in SR and 3.8\% in SPL over the strongest zero-shot baselines.

This significant performance improvement on MP3D highlights a critical advantage of our framework. While prior zero-shot methods often experience severe performance degradation when transitioning from HM3D to more visually and geometrically complex scenes due to their reliance on fixed semantic priors, our self-evolving agentic rule memory effectively mitigates this generalization gap. By continuously accumulating and dynamically retrieving experiential rules via a UCB-based mechanism, the agent adapts to unseen environments and corrects its own behaviors online without requiring any parameter updates. Furthermore, the consistently superior SPL across both benchmarks indicates that our method not only locates targets more reliably but also generates highly efficient, purposeful navigation trajectories. Rather than relying on post-hoc corrections after irreversible actions, our memory-guided preflection module empowers the foundation model to forecast failure risks and filter out low-value candidate frontiers prior to physical execution. This risk avoidance successfully prevents the exhaustive and redundant exploration loops that frequently constrain conventional zero-shot implementations, significantly reducing wasteful physical trial-and-error steps.


\subsection{Ablation Studies}

\begin{table}[!t]
\centering
\caption{Ablation studies on core modules. We ablate the Preflection, Memory-Evolving modules to validate their effectiveness on the HM3D and MP3D datasets. }
\vspace{-0.2cm}
\label{tbl:abl_module}
\resizebox{1\linewidth}{!}{
\begin{tabular}{c|cc|cc|cc}
\toprule 
\multirow{2.5}{*}{\textbf{Setting}} & \multicolumn{2}{c|}{\textbf{Module}}                        & \multicolumn{2}{c|}{\textbf{HM3D}}                         & \multicolumn{2}{c}{\textbf{MP3D}}  \\ 
\cmidrule(lr){2-3} \cmidrule(lr){4-5} \cmidrule(lr){6-7}
& \textbf{Preflection} & \textbf{Memory-Evolve} & \textbf{SR} $\uparrow$ & \textbf{SPL} $\uparrow$ & \textbf{SR} $\uparrow$  & \textbf{SPL} $\uparrow$ \\ \midrule 

1 & \multicolumn{2}{c|}{ }  & 64.7 & 32.5 & 43.9 & 15.8               \\
2 & \ding{51}  & ~          & 66.5 & 33.5 & 47.4 & 18.4              \\ 
3 & ~          & \ding{51}  & 66.7 & 33.6 & 48.3 & 18.7           \\
\midrule
4 & \ding{51}  & \ding{51}  & \textbf{67.3} & \textbf{33.9} & \textbf{49.0} & \textbf{19.1}              \\

\bottomrule 
\end{tabular}
}
\vspace{-0.5cm}
\end{table}

\begin{figure*}
    \centering
    \includegraphics[width=1.0\linewidth]{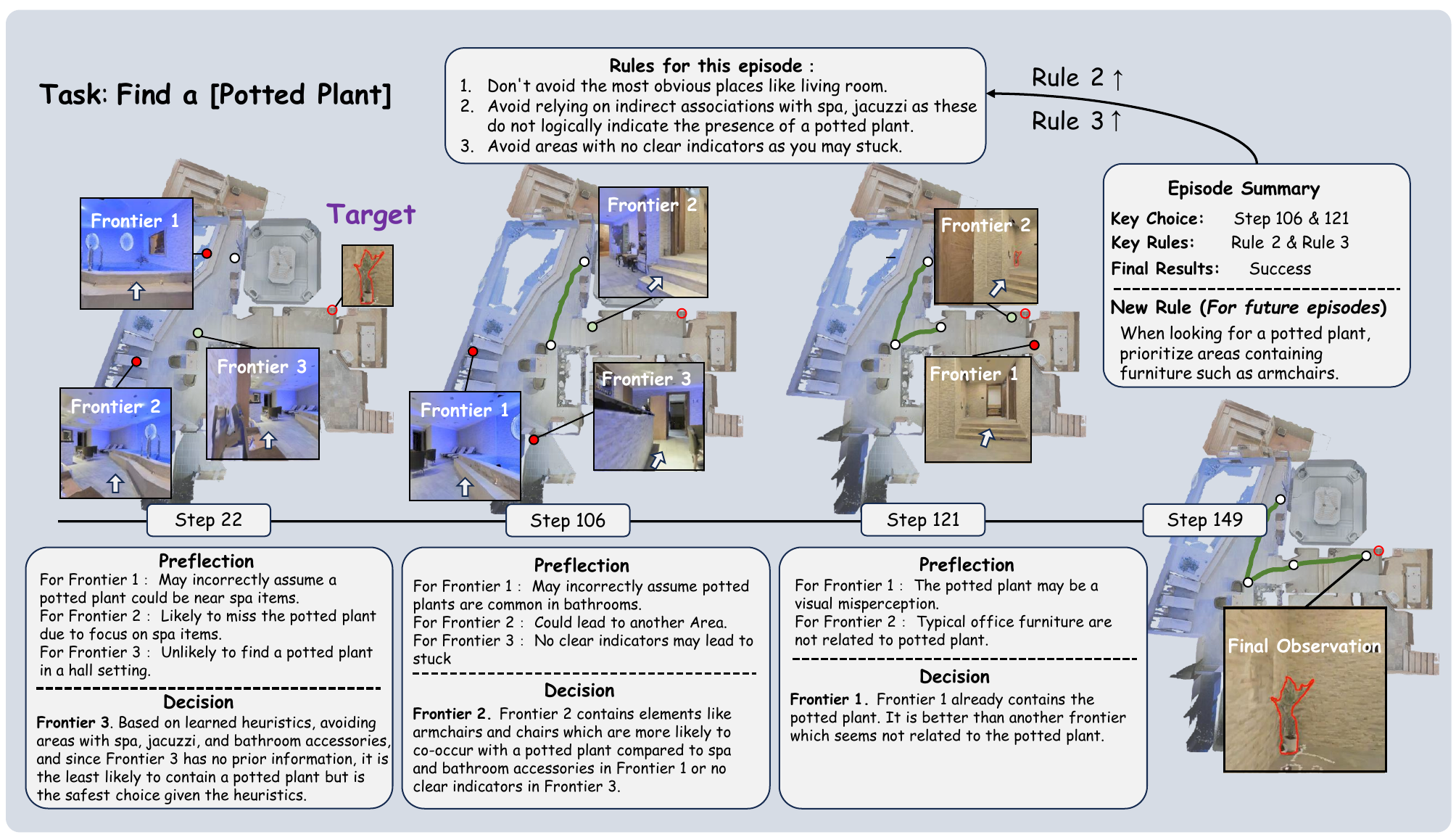}
    \vspace{-0.4cm}
    \caption{Navigation process visualization of \projecttitle. We provide the agent's trajectory alongside its reasoning process. During the episode, the agent utilizes Preflection to evaluate frontiers and proactively avoid unpromising rooms. After the episode, \projecttitle generates a new rule from the history and updates its Rule Bank. It illustrates how \projecttitle autonomously distills long-term navigation knowledge from short-term trial and error.}
    \label{fig: case study}
    \vspace{-0.1cm}
\end{figure*}

\subsubsection{Effectiveness of Preflection.} To verify the effectiveness of the proposed preflection, we disable it during inference. As shown in Table~\ref{tbl:abl_module}, integrating the memory-guided preflection module (Setting 2) yields a noticeable performance boost over the baseline (Setting 1), improving the SR by 0.8\% on HM3D and 3.5\% on the more challenging MP3D dataset. Notably, we also observe a steady increase in Success weighted by SPL. This validates our core hypothesis that shifting from reactive, post-hoc correction to prior-to-action anticipation significantly mitigates the prohibitive costs of physical trial-and-error. By forcing the LLM to explicitly predict frontier failure risks before committing to irreversible physical actions, the agent effectively filters out deceptive local optima and avoids dead ends. This proactive risk avoidance allows the agent to conserve its limited step budget, resulting in the generation of more purposeful and highly efficient navigation trajectories.

\subsubsection{Effectiveness of Memory and Evolving.} To evaluate the impact of continuous online experience accumulation, we isolate the self-evolving agentic rule memory alongside its UCB-based weight updating mechanism (Setting 3 in Table~\ref{tbl:abl_module}). Compared to the static baseline, empowering the agent to formulate and retrieve rules without explicit preflective reasoning still achieves an absolute SR improvement of 2.0\% on HM3D and a substantial 4.5\% on MP3D. The significant performance gain on the MP3D dataset highlights a critical bottleneck in existing zero-shot paradigms: the widely used static priors heavily relied upon by foundation models frequently fail in geometrically complex or unusual real-world environments. By dynamically extracting actionable rules from historical successes and failures via semantic-driven credit assignment, the agent successfully bridges this generalization gap, continuously adapting its behavior without requiring any parameter updates.

\begin{table}[t]
\centering
\caption{Ablation studies on LLM backbone. We investigate the impact of different LLM backbones within the Instruction Interpolator. The vision model is fixed to BLIP-2, and performance is evaluated on the HM3D and MP3D datasets.}
\label{tab:abla_large_model}
\vspace{-0.0cm}
\resizebox{\linewidth}{!}{
\begin{tabular}{cc|cc|cc}
\toprule
\multicolumn{2}{c|}{\textbf{Instruction Interpolator}} & \multicolumn{2}{c|}{\textbf{HM3D}} & \multicolumn{2}{c}{\textbf{MP3D}} \\
\cmidrule(lr){1-2} \cmidrule(lr){3-4} \cmidrule(lr){5-6}
\textbf{Vision} & \textbf{Language} & \textbf{SR} $\uparrow$ & \textbf{SPL} $\uparrow$ & \textbf{SR} $\uparrow$ & \textbf{SPL} $\uparrow$ \\
\midrule
\multirow{3}{*}{BLIP-2} & Qwen2.5-7B & $67.0$ & $33.9$ & $48.8$ & $19.0$ \\
 & Qwen3-8B   & $\mathbf{67.3}$ & $33.9$ & $\mathbf{49.0}$ & $\mathbf{19.1}$ \\
 & Qwen3.5-9B & $67.2$ & $\mathbf{34.1}$ & $\mathbf{48.9}$ & $\mathbf{19.2}$ \\
\bottomrule
\end{tabular}}
\vspace{-0.0cm}
\end{table}

\begin{figure*}
    \centering
    \includegraphics[width=1.0\linewidth]{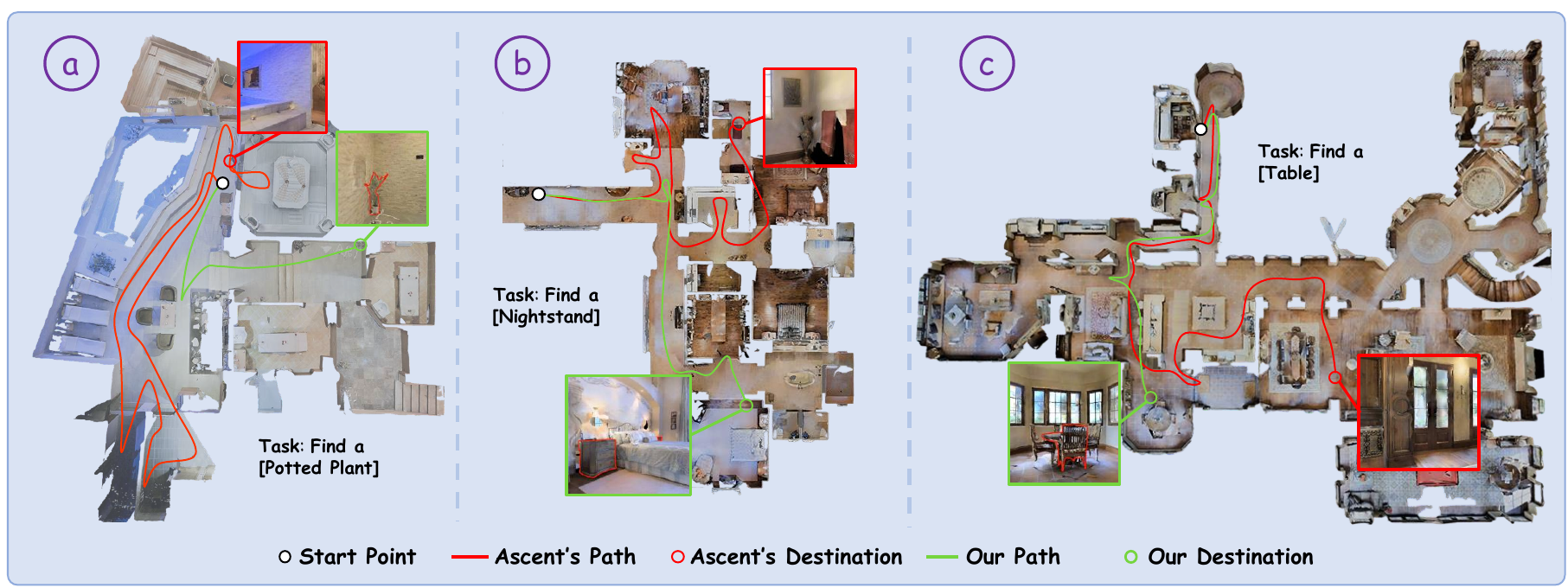}
    \vspace{-0.4cm}
    \caption{Qualitative Comparison. We visualize the start points, paths and destinations of \projecttitle and ASCENT~\cite{gong2026ascent} on several MP3D scenes. Across the three illustrated scenarios, the Ascent's trajectory is notably more convoluted. It frequently gets distracted by adjacent small rooms and narrow corridors, expending a significant number of steps to escape from these localized traps. In contrast, the paths of our proposed \projecttitle demonstrate better performance. This indicates \projecttitle achieves more efficient and robust results.}
    \vspace{-0.3cm}
    \label{fig: case study2}
\end{figure*}

\subsubsection{Effectiveness of LLM Backbone.}
To investigate the influence of the language foundation model on the overall navigation performance, we conduct an ablation study by changing the LLM backbone within the Instruction Interpolator, while keeping the vision foundation model fixed to BLIP-2. As presented in Table~\ref{tab:abla_large_model}, we evaluate the framework using Qwen2.5-7B, Qwen3-8B, and Qwen3.5-9B architectures. Notably, the empirical results indicate that \projecttitle exhibits limited sensitivity to the scale and specific iteration of the underlying LLM backbone. Across both datasets, the performance variance is practically negligible. For instance, transitioning from the 7B to the 9B parameter model yields only marginal fluctuations, maintaining a robust SR of approximately 67.0\%-67.3\% on HM3D and 48.8\%-49.0\% on the MP3D benchmark. We attribute this stability to the core architectural design of our framework. This suggests that in OGN tasks, stronger language foundation models may not lead to better results. Meanwhile, \projecttitle explicitly grounds the agent's decision-making in a dynamically maintained, self-evolving rule memory. During the preflection phase, the LLM is injected with structured local observations alongside high-value historical rules retrieved via the UCB selection criterion.



\subsection{Qualitative Evaluation}
To intuitively demonstrate the decision-making ability of \projecttitle, we provide a qualitative case study from the MP3D dataset, as shown in Figure \ref{fig: case study}. The agent is tasked with finding a <Potted Plant> in a multi-room layout. Instead of greedily exploring the nearest frontiers, \projecttitle demonstrates strong proactive risk avoidance. For instance, at Step 22, the agent triggers the preflection module before physical movement. By retrieving historical rules from the Rule Bank (e.g., Rule 2: "Avoid relying on indirect associations with spa, jacuzzi"), the LLM anticipates the failure risk of Frontier 1 and successfully bypasses it. Later at Step 106, preflection correctly guides the agent to prioritize an area with armchairs, anticipating a higher co-occurrence probability with the target. After the episode ends, \projecttitle continuously learns from its physical interactions. Following the trajectory, \projecttitle dynamically increases the UCB weights for the successful priors Rules 2 and 3. Furthermore, the LLM distills a new heuristic from this experience: "When looking for a potted plant, prioritize areas containing furniture such as armchairs." enriching the Rule Bank to guide future cross-scene explorations.

To further demonstrate the superiority of our proposed \projecttitle, we visualize and compare the navigation trajectories across several indoor scenarios from the MP3D dataset against Ascent, the best baseline method. As illustrated in Figure \ref{fig: case study2}, we present three distinct Object Goal Navigation tasks: finding a \textit{Potted Plant} (scene a), a \textit{Nightstand} (scene b) and a \textit{Table} (scene c). Across the three illustrated scenarios, Ascent's trajectory is notably more complex than that of our method. The red path frequently gets distracted by adjacent small rooms and narrow corridors, expending a significant number of steps to escape from these localized traps. In contrast, our method actively observes and reorients at the entrances of narrow passages, proactively preventing the agent from getting stuck. By leveraging the evolved rules and the preflection module, our agent is capable of planning a more direct, globally optimal route. For instance, in the layout of scene (c), \projecttitle bypasses irrelevant areas entirely and heads straight toward the dining area to locate the table. In conclusion, these qualitative visualizations corroborate our quantitative findings, demonstrating that \projecttitle effectively minimizes redundant exploration and exhaustive trial-and-error, thereby achieving significantly more efficient and goal-directed navigation in unseen environments.

\section{Discussion}
\subsection{Future Work} We plan to extend the proposed \projecttitle from a single-agent paradigm to a collaborative multi-agent navigation system. While our current self-evolving memory and preflection mechanisms significantly mitigate the physical trial-and-error costs for an individual agent, navigating large-scale and complex environments often necessitates distributed coordination.

\subsection{Limitations}
Despite its robust performance, our framework presents two primary limitations. First, it relies on 2D visual foundation models. Consequently, persistent 2D detection errors can directly lead to navigation failures. In future work, this could be mitigated by multi-view verification or by integrating active perception policies to enhance 3D spatial awareness. Second, the framework is potentially sensitive to unconventional layouts (e.g., a bed in a kitchen). Such environments may trap the agent in inefficient exploration loops. Addressing this issue by validating the rules or refining the update strategy requires further research.

\section{Conclusion}

We propose \projecttitle, a self-evolving, training-free framework that addresses the limitations of static priors in ZS-OGN. Our approach introduces a self-evolving rule memory that extracts actionable knowledge from past trajectories, utilizing a UCB-based retrieval strategy to balance semantic relevance with historical success. To minimize physical trial-and-error, we develop a memory-guided preflection module that shifts the exploration paradigm from passive correction to proactive risk avoidance. Extensive experiments on HM3D and MP3D demonstrate that \projecttitle significantly outperforms state-of-the-art zero-shot baselines.
\bibliographystyle{ACM-Reference-Format}
\bibliography{main}










\end{document}